\documentclass[runningheads]{llncs}
\usepackage{graphicx}
\usepackage{amsmath}
\begin{document}
\title{A privacy-preserving, distributed and cooperative FCM-based learning approach for Cancer Research}
%
\titlerunning{A privacy-preserving, distributed and cooperative FCM for Cancer Research}
%
\author{Jose L. Salmeron\inst{1,2} \and
Irina Ar\'evalo\inst{1}}
\authorrunning{Salmeron and Ar\'evalo}
%
\institute{Universidad Pablo de Olavide, Km. 1 Utrera road, 43013 Seville, Spain \and
Tessella, Altran World-Class Center for Analytics, c/ Campezo 1, 28022 Madrid, Spain
\email{salmeron@acm.org}\\
\email{iarebar@alu.upo.es}}
\maketitle              
\begin{abstract}
Distributed Artificial Intelligence is attracting interest day by day. In this paper, the authors introduce an innovative methodology for distributed learning of Particle Swarm Optimization-based Fuzzy Cognitive Maps in a privacy-preserving way. The authors design a training scheme for collaborative FCM learning that offers data privacy compliant with the current regulation. This method is applied to a cancer detection problem, proving that the performance of the model is improved by the Federated Learning process, and obtaining similar results to the ones that can be found in the literature. 

\end{abstract}

\section{Introduction}
\footnote{Preprint - Published in Rough Sets: International Joint Conference, IJCRS 2020}
Distributed Artificial Intelligence is a subfield of Artificial Intelligence that studies the coordination among several semi-autonomous agents called participants. Such systems are able to solve more complex problems involving a large amount of data, but there are privacy concerns about sharing sensitive information. 

Federated Learning is a novel approach to Distributed Artificial Intelligence that enables privacy-preserving communications by sharing the model (or gradients) instead of the data. A central server sends a model to be trained by the participants with their local data, who send the parameters of the model back to the server to be aggregated. After iterating this process, the output is a model that has been trained with the private information of all participants. 

This method is especially useful when dealing with sensitive data, from domains such as finance or healthcare. In this paper, the authors propose a Federated Fuzzy Cognitive Map approach to help diagnose malignant breast tumor cells.

The contributions of this paper can be summarized as follows:
\begin{itemize}
    \item Distributed learning. The authors propose a PSO-based FCM learning in a distributed way. 
    \item Privacy-preserving machine learning. The authors design a training scheme for collaborative FCM learning that offers data privacy. This proposal enables multiple participants to learn a FCM model on their own inputs, preserving the privacy of their own data and complying with data privacy regulations.
    \item Implementation. The authors evaluate the performance of the proposal with a well-known dataset of cancer diagnosis. The experimental results show that the proposal achieve a similar performance to other non-distributed methods and improves the performance of the non-collaborative approach. 
\end{itemize}

The rest of this paper is organized as follows. We discuss existing fundamentals of FCM and the learning approach in Section 2. Distributed Artificial Intelligence is described in Section 3. Then, we present the methodological proposal in Section 4. Section 5 describes the details of the experimental approach and the results. Finally, we draw a conclusion in Section 6.

\section{Fuzzy Cognitive Maps}

\subsection{Fundamentals}

Fuzzy Cognitive Maps (FCMs) were initially proposed by Kosko \cite{kosko.1986}. FCMs represent concepts, variables or features as nodes, the relationships between them as arcs, and the strengths of those relations as weights. It means that a weight assesses how much node $X$ causes node $Y$. The fuzzy weights for arcs are normalised on the range $\{[0,+1]|[-1,+1]\}$, depending if it includes negative values or not. The maximum negative influence is $- 1$ and the maximum positive influence is $+1$. The value zero shows that there is no relationship between the concepts. For computational purposes, FCMs can be described via a weight matrix (connection or adjacency matrix) which contains all weight values of edges between the concepts.

The relationships between the nodes are expressed by their weights. That is, if there is a positive causality between two nodes, then $\varpi_{ij}>0$. If there is a negative causality,then $\varpi_{ij}< 0$ and if there is no relationship between the two nodes, then $\varpi_{ij}=0$. The state of the nodes together is shown in the state vector $c =[c_1,c_2,\ldots,c_N]$ that gives a snapshot of nodes at any point of the instant in the scenario.

From a formal point of view, it is possible to represent a FCM as a 4-tuple $\Phi = \langle c, \mathcal{W}, f, r \rangle$, where $c=\{c_i\}_{i=1}^n$ is the state of the nodes with $n$ as the number of nodes, $\mathcal{W} = [\varpi_{ij}]_{n\times n}$  is the adjacency matrix representing the weights between the nodes, $f$ is the activation function, and $r$ is the nodes' range.

FCMs are dynamical systems involving feedback, where the effect of change in the state of a node may affect the state of other nodes, which in turn can affect the former node \cite{napoles.2020}.

The dynamic starts with an initial vector state $c(0)=\big(c_1(0),\ldots, c_n(0)\big)$, which represents the initial state (value) of each node. The new state of the nodes is computed as an iterative process. It includes an activation function \cite{bueno.2009} for mapping monotonically the node state into a normalized range $\{[0, +1]|[-1, +1]\}$. If the range is $[0, +1]$, the unipolar sigmoid is the most used activation function, but hyperbolic tangent is the most used when the range is $[-1, +1]$.

The component $i$ of the vector state at time $t,$ $c_i(t),$ can be computed as

\begin{equation}
c_i(t) = f\Bigg(\sum_{j=1}^n \varpi_{ji}\cdot c_j(t-1)\Bigg).
\end{equation}

Some systems include nodes whose states should be steady because their states are not related with the dynamics of the system but their state has some influence on the state of the other nodes (i.e. sun radiation, wind speed and so on). In such cases, the state of the node is the same along the dynamics $c_i(t)=c(t-1)\;|\; c_i\in \mathcal{O},$ where $\mathcal{O}$ is the set of output concepts.

If the activation function $f$ is unipolar sigmoid, then the component $i$ of the vector state $c_i(t)$ at the instant $t$ is computed as follows

\begin{equation}
    c_i(t) = \Big(1+e^{-\lambda\cdot \sum_{j=1}^n \varpi_{ji}\cdot c_j(t-1)}\Big)^{-1}
\end{equation}

If the activation function $f$ is hyperbolic tangent, then the component $i$ of the vector state $c_i(t)$ at the instant $t$ is computed as follows

\begin{equation}
    c_i(t) = \frac{e^{\lambda\cdot\sum_{j=1}^n \varpi_{ji}\cdot c_j(t-1)}-e^{-\lambda\cdot\sum_{j=1}^n \varpi_{ji}\cdot c_j(t-1)}}{e^{\lambda\cdot\sum_{j=1}^n \varpi_{ji}\cdot c_j(t-1)}+e^{-\lambda\cdot\sum_{j=1}^n \varpi_{ji}\cdot c_j(t-1)}}
\end{equation}

After the dynamics, the FCM reaches one of the three following states after a number of iterations: it settles down to either a fixed pattern of node values (the so-called hidden pattern), a limited cycle, or a fixed-point attractor.

\subsection{Augmented FCMs}
\label{afcm}
According to the FCM literature \cite{lopez.2014}, an augmented adjacency matrix is built by aggregating the adjacency matrix of each FCM. The elements' aggregation depends on whether there are common nodes. If the adjacency matrices had no common nodes, the elements $\varpi_{ij}$ in the augmented matrix ($\displaystyle\otimes_{i=1}^{N}$) are computed by adding the adjacency matrix of each FCM model ($\mathcal{W}_{i}$). 

The addition method when the adjacency matrices have not common nodes is known as direct sum of matrices, and the augmented matrix is denoted as $\otimes_{i=1}^{N}\varpi_{i}$. Given a couple of FCMs with no common nodes and even different number of nodes with adjacency matrices $[\varpi_{ij}]_{n\times n}$ and $[\varpi_{kl}]_{m\times m}$, the resulting augmented adjacency matrix is as follows

\begin{equation}
\begin{array}{rl}
\displaystyle\otimes_{i=1}^{N}\varpi_{i} &= \textrm{\textbf{diag}}(\varpi_{jk}, \varpi_{lo}) \\[0.25cm]
	&=
		\left(
			\begin{array}{cc}
				 0 &  [\varpi_{jk}]_{r\times r}  \\
				{[\varpi_{lo}]}_{m\times m}  & 0
			\end{array}
		\right)
	\end{array}
\end{equation}

\noindent where $N$ is the number or adjacency matrices to add, zeroes are actually zero matrices and the dimension of $\displaystyle\otimes_{i=1}^{N}\varpi_{i}$ is $[\cdot]_{m+r\times m+r}$. In the case of common nodes, they would be computed as the average or weighted average of the states of the nodes in each adjacency matrix.

\subsection{FCM for classification}
FCMs classification capabilities have been analysed by \cite{papakostas.2010}. In general terms, the main goal of a conventional classifier is the mapping of an input to a specific output according to a pattern. In this proposal, the input concepts represent the features of the dataset, while the output concepts are the classes’ labels where the patterns belong. 

\begin{figure}
\centerline{\includegraphics[width=5.0cm]{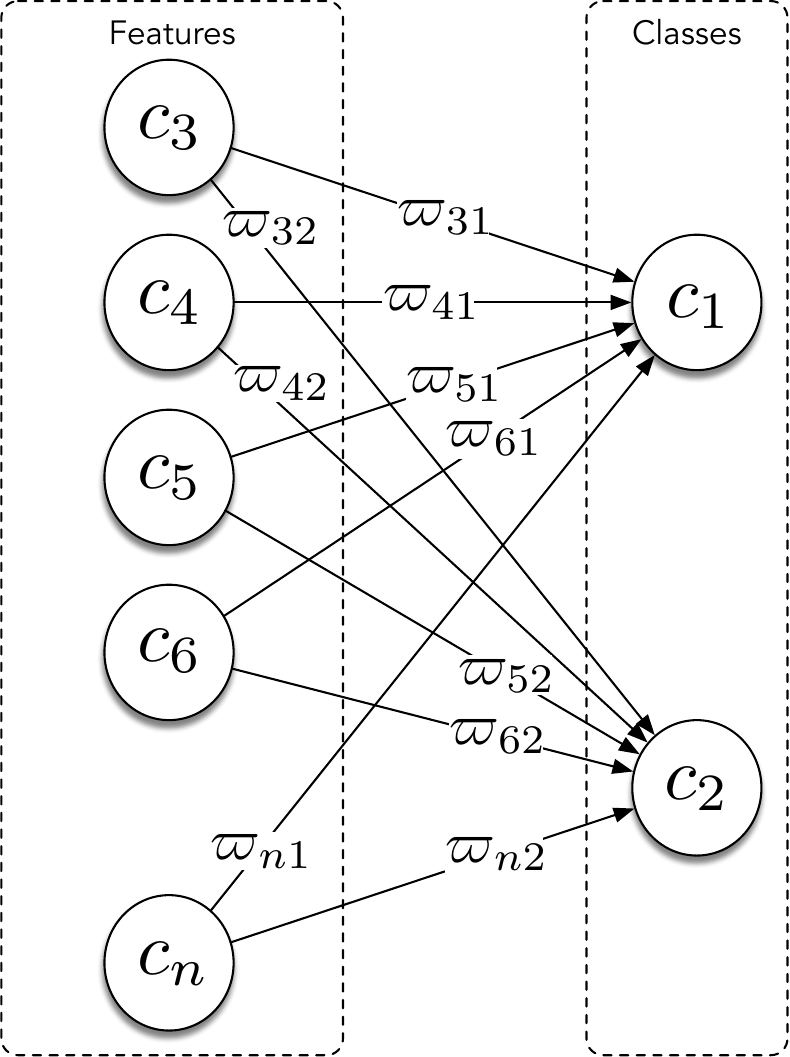}}
\caption{Fuzzy Cognitive Maps classifier} \label{fcmcla}
\end{figure}

Figure \ref{fcmcla} shows the typical topology of a FCM classifier where the state of the concepts $c_{1}$ and $c_{2}$ defines the class where the input vector state belongs. In that sense, if $c_{1}>c_{2}$ the input vector state belongs to class 1 but if  $c_{1}<c_{2}$ the input vector state belongs to class 2. Note that $c_{i}\in\{[-1,+1],[0,+1]\}$, therefore if $c_{1}=0.03$ and $c_{2}=0.1$, then the input vector state belong to class 2.

\subsection{PSO-based FCM learning}
FCM learning endeavours are commonly focused on building the adjacency matrix based either on the available historical raw data or on expert knowledge. FCM learning approaches could be divided into three categories \cite{salmeron.2019}: Hebbian, population-based, and hybrid, mixing the main aspects of Hebbian-based and population-based learning algorithm.

The goal of the Hebbian-based FCM learning approaches is to modify adjacency matrices leading the FCM model to either achieve a steady state or converge into an acceptable region for the target system.

Population-based approaches do not need the human intervention. They compute adjacency matrices from historical raw data that best fit the sequence of input state vectors (the instances of the dataset). The learning goal of FCM evolutionary learning is to generate optimal adjacency matrix for modeling systems behaviour.

In this sense, Salmeron et al. \cite{salmeron.2017} proposed an advanced decision support tool based on consultations with a group of experienced medical professionals using FCMs trained with Particle Swarm Optimization (PSO). Also, Salmeron and Froelich \cite{salmeron.2016} apply PSO for time series forecasting.

PSO is a bio-inspired, population-based and stochastic optimization algorithm. The PSO algorithm generates a swarm of particles moving in an $n$-dimensional search space which must include all potential candidate solutions.

In order to train the FCM adjacency matrices we take into account the $k^{th}$ particle's position (a candidate solution or adjacency matrix), denoted as $\varpi_k=(\varpi_{k_1},\ldots,\varpi_{k_j})$ and its velocity, $v_k=(v_{k_1},\ldots,v_{k_j})$. Note that each particle is a potential solution or FCM candidate and its position $\varpi_k$ represents its adjacency matrix.

Each particle's velocity and position are updated at each time step. The position and the velocity of each particle is computed as follows

\begin{subequations}
    \begin{align}
         \varpi_k(t+1) &= \varpi_k(t) + v_k(t)\\
         v_k(t+1) & = v_k(t) + U(0,\phi_1)\otimes (\dot{\varpi}_k -\varpi_k(t)) + U(0,\phi_2)\otimes (\ddot{\varpi}_k -\varpi_k(t))
    \end{align}
\end{subequations}

\noindent where $U(0,\phi_i)$ is a vector of random numbers generated from a uniform distribution within $[0,\phi_i],$ generated at each iteration and for each particle. Also, $\dot{\varpi}_k$  is the best position of particle $k$ in all former iterations and $\ddot{\varpi}_k$ the best position of the whole population in all previous iterations and $\otimes$ is the component-wise multiplication.

The PSO algorithm's goal is to locate all the particles in the global optima to a multidimensional hyper-volume. The fitness function used in this research is the complement of the Jaccard similarity coefficient ($\overline{J}=(Y\times \hat{Y})\setminus J$). The Jaccard score computes the average of Jaccard similarity coefficients between pairs of the sets of labels. The Jaccard similarity coefficient of the $i$-th samples, with a ground truth label set and a predicted label set. The complement operation is needed in terms of minimization of the fitness function. The Jaccard similarity coefficient's complement is computed as follows

\begin{equation}
\overline{J}(y_{i},\hat{y}_{i}) = 1- \frac{\vert y_{i} \cap\hat{y}_{i}\vert}{\vert y_{i} \cup\hat{y}_{i}\vert}
\end{equation}

The fitness function is sampled after each particle position update and is the objective function used to compute how close a given particle is in order to be able to achieve the set aims. 

\section{Methodological proposal}
\subsection{Fundamentals}
Distributed Artificial Intelligence is a subset of Artificial Intelligence that allows the sharing of information among several agents or participants that interact by cooperation, by coexistence or by competition. Such system manages the distribution of tasks, being therefore more apt to solve complex problems, especially if they involve a large amount of data. 

One of the methods available to construct a distributed artificial intelligence system is Federated Learning, proposed by McMahan et al. \cite{mcmahan.2016} and further developed in Konecny et al. \cite{konecn.2016} and McMahan and Ramage \cite{mcmahan_ramage.2017}. In such system, a central server constructs a model, usually a neural network, and sends it to the participants, who train the model in their private data. Their data never leaves their local devices, therefore ensuring privacy and security. The parameters of the participant's model are then averaged to obtain a global model. This process may be iterated till convergence. 

\begin{figure}
\centerline{\includegraphics[width=9.0cm]{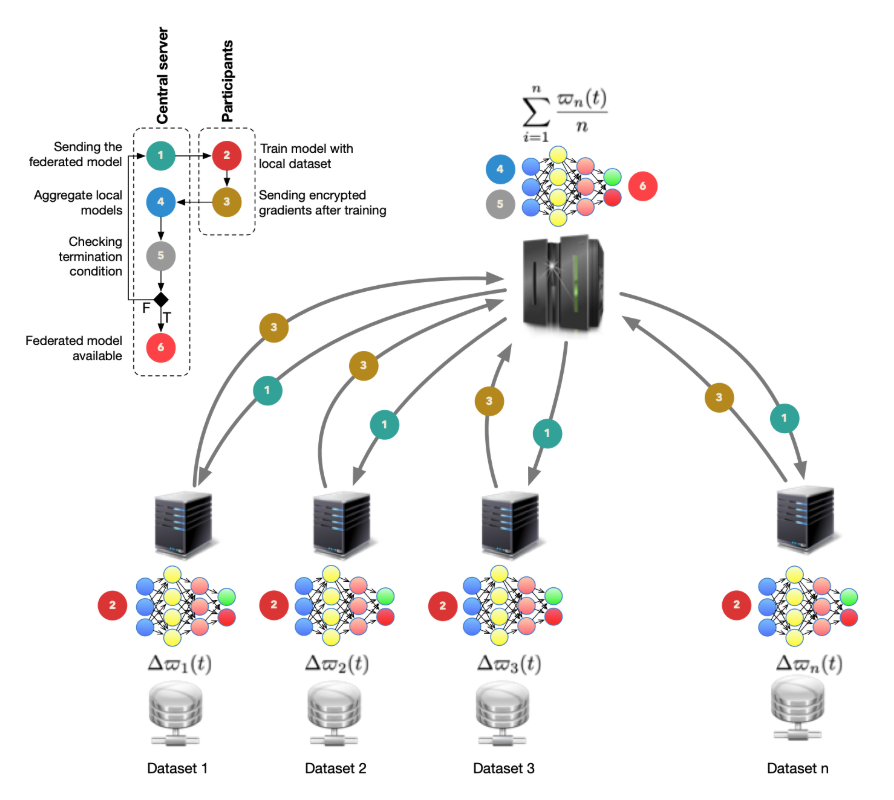}}
\caption{Federated Learning process} \label{FL}
\end{figure}

Described in a formal way, a Federated Learning project is composed by a central server and the participants. The central server is responsible for managing the federated model and the communications with the participants. The participants own the datasets and train the partial models. The whole process is described in Figure \ref{FL} and it is as follows:

\begin{enumerate}
\item The central server sends a federated model to each participant. If it is the initial iteration the federated model is proposed by the central server.
\item Each participant trains the received model with their own private dataset.
\item After the partial model is trained, each participant sends the parameters of the model or its gradients to the central server, encrypted to ensure privacy.
\item The central server aggregates the partial model and builds the federated model.
\item The central server checks the termination condition and if it is accomplished the federated model is finished, otherwise the process goes back to step 1.
\end{enumerate}

When the researchers at Google first defined Federated Learning, their initial idea was to allow Android mobile phones to collaborative construct a prediction model without migrating the training data from the phone (see McMahan et al. \cite{mcmahan_ramage.2017} from the Google AI Blog). A first application they had was to use FL in Gboard on Android, the Google Keyboard, which predicts the most probable next phrase or word based on the user-generated preceding text. Recently, Federated Learning has improved this process, allowing the use of more accurate models with lower latency, ensuring privacy and less power consumption. 

One of the main advantages of Federated Learning is the promise of secure and private distributed machine learning, but there are risks associated with sharing data among several agents, such as the reconstruction of training examples from the neural network parameters, the uploading of private data from the agents to the central server, and the protection of the models as intellectual property of the companies. 
There is a large research interest in privacy-preserving methods applied to Federated Learning, such as the application of Differential Privacy, Secure Multi-Party Computation or Homomorphic encryption. 

\subsection{FCM distributed learning}
\begin{figure}
\centerline{\includegraphics[width=8.7cm]{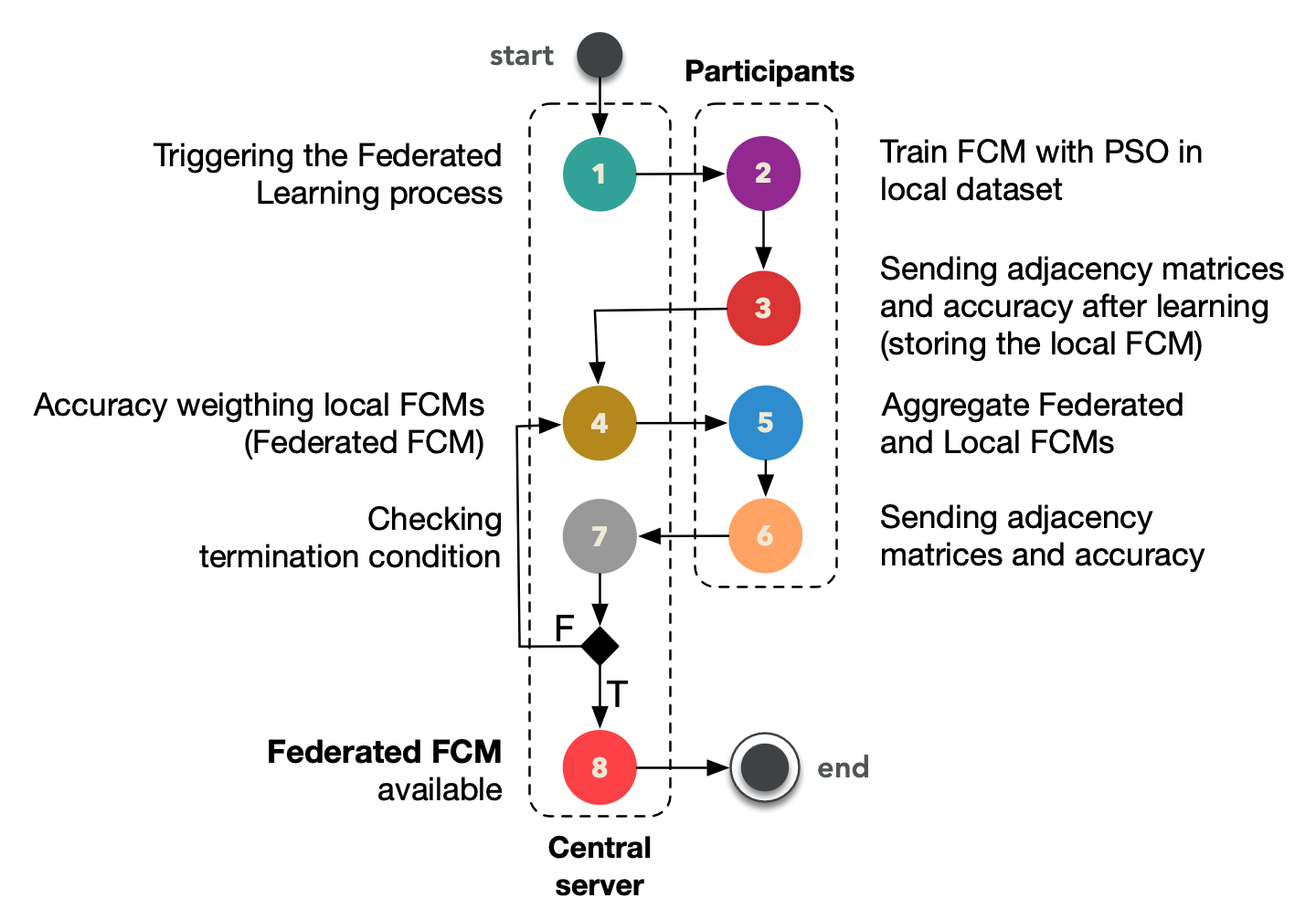}}
\caption{Proposed methodology} \label{proposal}
\end{figure}

The proposed methodology combines Federated Learning with learning FCMs using Particle Swarm Optimization. The process is shown in Figure \ref{proposal} and it is explained as follows.
\begin{enumerate}
    \item Triggering the Federated Learning process. The central server triggers the process in the participants machines.
    \item Training FCM in the local dataset. Each participant trains a local FCM with their own dataset. The authors apply PSO but this methodology is agnostic to the learning approach. The FCM dynamics is considered steady when the difference between two consecutive vector states is under $tol= 0.00001$
    \item Sending the trained adjacency matrices and local accuracy for this stage to the central server. The local FCM is stored in the participant devices.
    \item Weighting local FCMs using accuracy. The central server aggregates the local FCMs weighting by the accuracy. The aggregation method have been detailed as Section \ref{afcm}.
    \item Aggregating Federated and Local FCMs. The participants aggregate the Federated FCM from the central server and their own local FCM.
    \item Sending adjacency matrices and accuracy. Participants send again the local adjacency matrices and the new local accuracy.
    \item Checking termination condition. The central server checks if the Federated process has been run 20 iterations as termination condition. If it is not accomplished then it goes back to the step 4.
    \item If the termination condition is accomplished then a Federated FCM is achieved.
\end{enumerate}

The main contribution of this paper is the application of Federation Learning paradigm for privacy-preserving FCM distributed and coorperative learning.

\section{Experimental approach}
\subsection{Dataset}
Breast cancer is one of the most common cancers among women, accounting for 25\% of all cancer cases that affect women worldwide. According to the American Cancer Society, when breast cancer is detected early, and is in the localized stage, the 5-year relative survival rate is 99\%, which makes the early diagnosis of breast cancer a main key in the prognosis and chance of survival of such types of cancer. 

In recent years the use of Machine Learning algorithms in medicine has increased exponentially, with applications such as EEG analysis and Cancer detection. For example, automatized algorithms have been use to examine biological data such as DNA methylation and RNA sequencing to infer which genes can cause cancer and which genes can instead be able to suppress its expression. 

In this paper the authors will use the Breast Cancer Wisconsin Dataset, created by Dr. William H. Wolberg, physician at the University Of Wisconsin Hospital at Madison, and made publicly available at the UC Irvine Machine Learning Repository. The dataset comprises data from digitized images of the fine-needle aspirate of a breast mass that describes features of the nucleus of the current image of 569 patients, of which 212 are malignant and 357 are benign cases. 

The first two features correspond to the identifier number and the diagnosis status (our target). The remaining attributes are thirty real attributes that measure the mean, the standard error, and the worst radius, texture, perimeter, area, smoothness, compactness, concave points, concavity, symmetry, and fractal dimension of the nucleus of the solid breast mass 
These data were obtained using a graphical computer program called Xcyt, which is capable of perform the analysis of cytological features based on a digital scan. More details can be found in \cite{street.1993}, \cite{mangasarian.1995}.

\subsection{Results}

After 20 iterations of the Federated Learning process, the Fuzzy Cognitive Map-based classifier is able to predict whether the tumor is malignant with an average accuracy of 0.9383 across all participants, improving the accuracy of a single Fuzzy Cognitive Map trained in the whole data, and the accuracy in each participant before the federation. 

The goal of this paper is not the accuracy of the proposal but a distributed and privacy-preserving approach. Nevertheless, our results are similar to the ones found in literature \cite{wang.2020}. 

\begin{table}
\centering
\caption{Results of the experiments}\label{results}
\begin{tabular}{|c| c| c|}
\hline
 & \textbf{Accuracy} &         \textbf{Accuracy}\\
 \textbf{Participant}& \textbf{pre-Federated} &         \textbf{post-Federated}\\
 & \textbf{ Learning} &         \textbf{Learning}\\
\hline
1 &  0.7727  & 0.9091 \\
2 & 0.9130 & 0.9130 \\
3 & 0.8696 & 0.8696 \\
4 & 0.9565 & 1.0000 \\
5 & 1.0000 & 1.0000 \\
\hline
\end{tabular}
\end{table}

\section{Conclusions}

This paper proposes an innovative methodology for learning Fuzzy Cognitive Maps with Federated Learning. It is a step forward for Distributed Artificial Intelligence and accomplishes the privacy-preserving requirements of the society.

In addition, the authors have developed a method for distributed Fuzzy Cognitive Maps that improves the accuracy of both the algorithm trained in the whole dataset in a local node and the participant's algorithms before the Federated Learning process.

This method was applied to a cancer detection problem, obtaining an accuracy of 0.9383. The participants in this process do not share their private data, therefore forming a privacy-preserving distributed system.

%
%
%
%
\bibliographystyle{main}
\bibliography{main}

\begin{thebibliography}{10}
\providecommand{\url}[1]{\texttt{#1}}
\providecommand{\urlprefix}{URL }
\providecommand{\doi}[1]{https://doi.org/#1}

\bibitem{bueno.2009}
Bueno, S., Salmeron, J.L.: Benchmarking main activation functions in fuzzy
  cognitive maps. Expert Systems with Applications  \textbf{36}(3 Part 1),
  258--268 (2009)

\bibitem{konecn.2016}
Konecn{\'y}, J., McMahan, B., Ramage, D., Richt{\'a}rik, P.: Federated
  optimization: Distributed machine learning for on-device intelligence. ArXiv
  \textbf{abs/1610.02527} (2016)

\bibitem{kosko.1986}
Kosko, B.: Fuzzy cognitive maps. International Journal of Man-Machine Studies
  \textbf{24}(1),  65--75 (1986)

\bibitem{lopez.2014}
Lopez, C., Salmeron, J.L.: Modeling maintenance projects risk effects on erp
  performance. Computer Standards \& Interfaces  \textbf{36}(3),  545--553
  (2014)

\bibitem{mcmahan.2016}
McMahan, B., Moore, E., Ramage, D., y~Arcas, B.A.: Federated learning of deep
  networks using model averaging. ArXiv  \textbf{abs/1602.05629} (2016)

\bibitem{mcmahan_ramage.2017}
McMahan, B., Ramage, D.: Google ai blog (Apr 2017),
  \url{https://ai.googleblog.com/2017/04/federated-learning-collaborative.html}

\bibitem{napoles.2020}
N\'apoles, G., Jastrzebska, A., Mosquera, C., Vanhoof, K., Homenda, W.:
  Deterministic learning of hybrid fuzzy cognitive maps and network reduction
  approaches. Neural Networks  \textbf{124},  258--268 (2020)

\bibitem{papakostas.2010}
Papakostas, G., Koulouriotis, D.: Fuzzy Cognitive Maps, chap. Classifying
  Patterns Using Fuzzy Cognitive Maps, pp. pp 291--306. Studies in Fuzziness
  and Soft Computing (volume 247), Springer (2010)

\bibitem{salmeron.2016}
Salmeron, J.L., Froelich, W.: Dynamic optimization of fuzzy cognitive maps for
  time series forecasting. Knowledge-Based Systems  \textbf{105},  29--37
  (2016)

\bibitem{salmeron.2019}
Salmeron, J.L., Mansouri, T., Moghadam, M.R.S., Mardani, A.: Learning fuzzy
  cognitive maps with modified asexual reproduction optimisation algorithm.
  Knowledge-Based Systems  \textbf{163},  723--735 (2019)

\bibitem{salmeron.2017}
Salmeron, J.L., Rahimi, S.A., Navali, A.M., Sadeghpour, A.: Medical diagnosis
  of rheumatoid arthritis using data driven pso--fcm with scarce datasets.
  Neurocomputing  \textbf{232},  65--75 (2017)

\bibitem{mangasarian.1995}
Street, W., Wolberg, W., Mangasarian, O.: Breast cancer diagnosis and prognosis
  via linear programming. Oper. Res.  \textbf{43}(4),  570--577 (Aug 1995).
  \doi{10.1287/opre.43.4.570}, \url{https://doi.org/10.1287/opre.43.4.570}

\bibitem{street.1993}
Street, W., Wolberg, W., Mangasarian, O.: Nuclear feature extraction for breast
  tumor diagnosis. vol.~1993 (01 1999). \doi{10.1117/12.148698}

\bibitem{wang.2020}
Wang, S., Wang, Y., Wang, D., Yin, Y., Wang, Y., Jin, Y.: An improved random
  forest-based rule extractionmethod for breast cancer diagnosis. Applied Soft
  Computing  \textbf{86} (2020)

\end{thebibliography}
\end{document}